\documentclass[letterpaper, 10 pt, conference]{ieeeconf}  

\IEEEoverridecommandlockouts                              

\usepackage{graphics, graphicx} 
\usepackage{amsmath} 
\usepackage{amssymb}  
\usepackage{siunitx}
\usepackage{algorithm}
\usepackage{algpseudocode}

\title{\LARGE \bf
Reinforcement Learning on Reconfigurable Hardware: Overcoming Material Variability in Laser Material Processing}

\author{Giulio Masinelli$^{1, 2}$, Chang Rajani$^{1}$, Patrik Hoffmann$^{1}$, Kilian Wasmer$^{1}$, and David Atienza$^{2}$ 
\thanks{$^{1}$Laboratory for Advanced Materials Processing (LAMP), Swiss Federal Laboratories for Materials Science and Technology (Empa), Thun, Switzerland. Correspondence to {\tt\small giulio.masinelli@empa.ch}}%
\thanks{$^{2}$Embedded Systems Laboratory (ESL), École Polytechnique Fédérale de Lausanne (EPFL), Lausanne, Switzerland}%
\thanks{© 2025 IEEE. Personal use of this material is permitted. Permission from IEEE must be obtained for all other uses, in any current or future media, including reprinting/republishing this material for advertising or promotional purposes, creating new collective works, for resale or redistribution to servers or lists, or reuse of any copyrighted component of this work in other works.}
}

\begin{document}

\maketitle
\thispagestyle{empty}
\pagestyle{empty}


\begin{abstract}
Ensuring consistent processing quality is challenging in laser processes due to varying material properties and surface conditions. Although some approaches have shown promise in solving this problem via automation, they often rely on predetermined targets or are limited to simulated environments. To address these shortcomings, we propose a novel real-time reinforcement learning approach for laser process control, implemented on a Field Programmable Gate Array to achieve real-time execution. Our experimental results from laser welding tests on stainless steel samples with a range of surface roughnesses validated the method's ability to adapt autonomously, without relying on reward engineering or prior setup information. Specifically, the algorithm learned the optimal power profile for each unique surface characteristic, demonstrating significant improvements over hand-engineered optimal constant power strategies --- up to 23\% better performance on rougher surfaces and 7\% on mixed surfaces. This approach represents a significant advancement in automating and optimizing laser processes, with potential applications across multiple industries.

\end{abstract}

\section{INTRODUCTION}

Laser material processing, including applications such as welding, cutting, and additive manufacturing, is a critical technology widely employed in various industrial sectors, such as automotive manufacturing \cite{Spottl2014}, aerospace engineering \cite{serrano2020laser}, and electronics assembly \cite{liu2017review}. These processes --- valued for their precision, speed, minimal mechanical interaction, and ability to produce high-quality results --- involve focusing a high-power laser beam onto the material's surface, creating localized effects such as melting, vaporization, or chemical reactions \cite{steen2010laser}. However, ensuring consistent processing quality across different materials and conditions poses significant challenges. For example, variations in material properties --- such as surface roughness, composition, and thickness --- can affect the process outcomes, necessitating real-time adjustments to the laser parameters.

\begin{figure}
    \centering\includegraphics[width=0.9\linewidth]{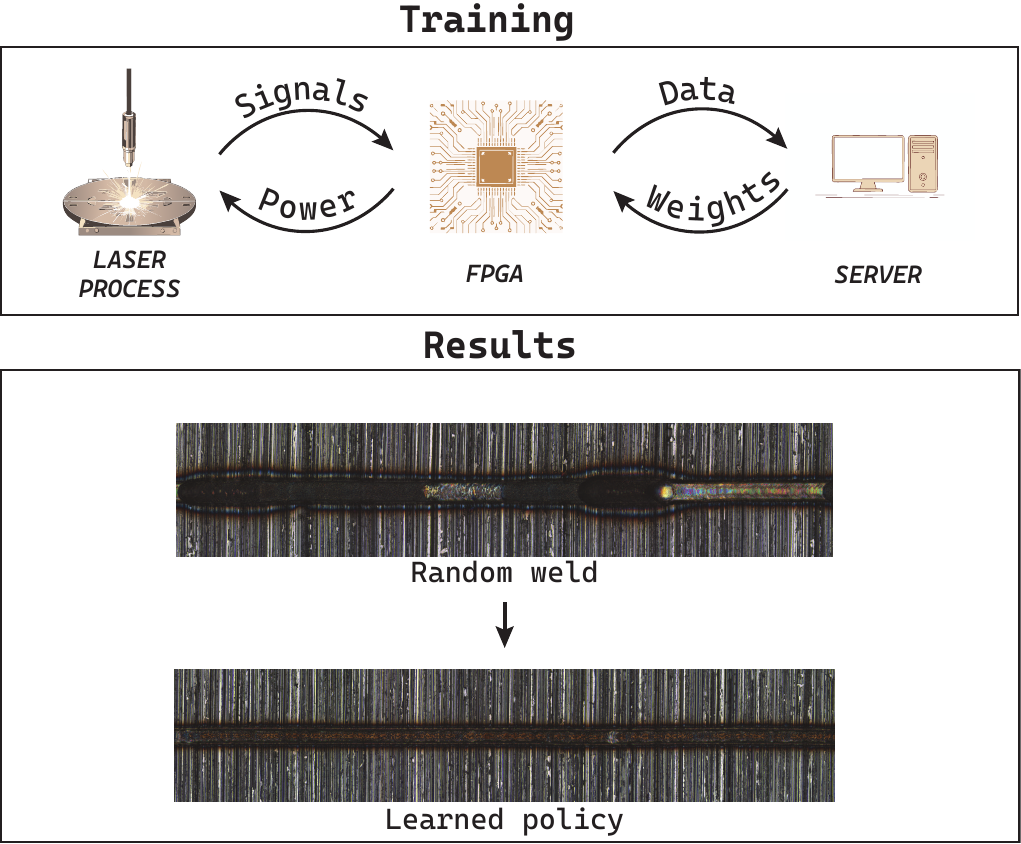}
    \vspace{-5pt}
    \caption{\footnotesize Illustration of the proposed method. \textbf{Top:} The FPGA receives optical signals from the process zone and uses its onboard policy network to determine the laser power in real-time. Between processing runs, the collected data is sent to a server where RL is used to train the policy. \textbf{Bottom:} The policy initially starts with random actions and learns to optimize the process outcome, achieving the best possible results while avoiding defects such as keyhole formation.}
    \label{fig:illustration}
    \vspace{-20pt}
\end{figure}

Traditionally, these adjustments are manually optimized by engineers, a process that is time-consuming and prone to errors, since even minor changes in material properties can require extensive reprogramming of the laser system \cite{Schmidt2006}. This manual approach is increasingly inadequate for meeting the demands of modern manufacturing, necessitating the development of advanced control strategies. Among these, Reinforcement Learning (RL) has emerged as a particularly promising technique, demonstrating success in complex real-world control tasks such as robotics \cite{gu2017deep}, nuclear fusion \cite{degrave2022magnetic}, and even laser welding \cite{masinelli2020adaptive, kaneko2023reinforcement}.

Specifically, RL methods learn control policies by interacting with the environment and receiving feedback, making them particularly suited for dynamic and complex processes like laser material processing. Nevertheless, although RL has shown considerable success in various industrial applications, its application to laser processes remains relatively unexplored. In fact, the fluctuating dynamics of these processes, combined with difficult-to-interpret sensor signals and challenging performance evaluations, present unique challenges for RL methods.

Despite these obstacles, several notable attempts have been made to apply RL to laser process control. For example, Günther et al. (2016) \cite{gunther2016intelligent} pioneered this approach in laser welding, using weld width as input and a sigmoid-transformed depth error as reward. Although innovative, their method relied on a predetermined reference depth and was primarily evaluated in a simplified simulation, potentially limiting its real-world applicability. Advancing this concept, Masinelli et al. (2020) \cite{masinelli2020adaptive} introduced a more sophisticated RL approach, incorporating multiple input signals, including optical and acoustic emissions, to control laser power. However, their method faced generalizability challenges due to a hand-engineered reward function based on a Machine Learning (ML) classifier, and its implementation on a commercial PC introduced latency and non-deterministic execution --- limiting its possibility to adapt to surface changes. More recently, Kaneko et al. (2023) \cite{kaneko2023reinforcement} proposed an RL control method for laser welding that minimizes bead width error by adjusting both laser power and scan speed. While promising, this approach required a predetermined target bead width and --- similar to the work of Günther et al. --- was only validated in a simulated environment.

These studies highlight the potential of RL in laser process control while underscoring persistent challenges in the field:
\begin{itemize}
    \item The need for a more generalizable approach rooted in physics-based understanding, rather than relying on situation-specific reward functions or predetermined targets.
    \item Reliance on simulated environments for validation, potentially overlooking real-world complexities.
    \item Implementation challenges that introduce latency or limit the system's ability to respond to rapid changes in processing conditions.
\end{itemize}

To address these limitations, we present a novel real-time method for laser material processing, utilizing reinforcement learning to adjust laser parameters based on optical signals from the process zone (PZ). Our system combines an FPGA-implemented policy for rapid execution with a server-based training component, leveraging the FPGA's high-speed data processing capabilities for real-time control while utilizing the server's computational power to continuously update the model's parameters.

Although our approach is general and applicable to various laser processes, we demonstrate its effectiveness in the context of laser welding. Specifically, our work advances the field through the following key contributions: 
\begin{itemize} 
\item \textbf{Real-Time Closed-Loop Control on FPGA}: We develop a closed-loop control system implemented on an FPGA, capable of adjusting laser power in real time at microsecond scales, ensuring rapid response to dynamic conditions. 
\item \textbf{Autonomous Adaptation Without Prior Knowledge}: Our RL method adapts to different surface conditions and material properties, without requiring assumptions or manual tuning. 
\item \textbf{Experimental Validation on Real-World Surfaces}: We validate our system in laser welding, conducting extensive experiments on 316L stainless steel samples with varying surface roughness, demonstrating significant improvements over optimal constant power strategies. 
\end{itemize}

\section{Experimental Methodology}\label{sec:methodology}

The core components of the setup include a laser source, an optical laser head, a stage holder mounted on a moving stage, and optical photodiodes.

The laser source is a fiber laser system LFS150 (Coherent Switzerland AG, Switzerland), with a maximum output capacity of 250 W at a wavelength of 1070 nm. The diameter of the laser spot measures 34 \si{\micro\meter} (within $1/e^2$) in the focal plane. Operating in continuous-wave (CW) mode, the source allows power modulation via an external voltage source (0--5 V). For this paper, we configured the system to operate within a power range of 25--100 W, which corresponds to the full range of the external voltage control.

All experiments were conducted in air at atmospheric pressure. To prevent weld oxidation and minimize plume absorption and scattering effects on optical signals, an argon flow was directed to the PZ blowing at a constant rate of 2 L/min.

Line welds were achieved by mounting samples on a linear Zaber LRT0250DL stage (Zaber Technologies, Canada), moving at a constant velocity of 50 mm/s. The movement of the sample was synchronized with the laser source to ensure that irradiation began only after the stage reached the set velocity. 

\subsection{Sensor System}

The laser head incorporates a customized optical system that directs the on-axis optical radiation from the PZ to two photodiodes, enabling simultaneous measurement of optical reflection and emission. For optical reflection (OR) measurements, we employed a silicon (Si) free-space amplified photodetector (PDA100A2, Thorlabs) with a spectral range of 320--1100 nm, set at a gain of 30 dB. Reflection data provide information on surface conditions and transitions between melting regimes, such as from conduction mode (where heat is transferred into the material with minimal vaporization) to keyhole mode (where intense laser power creates a vapor-filled cavity or `keyhole') \cite{Taherkhani2021}. For optical emission (OE) measurements, we used an Indium Gallium Arsenide (InGaAs) free-space amplified photodetector (PDA10CS2, Thorlabs) with a spectral range of 900--1700 nm, set at 50 dB gain. This detector was paired with an NF1064-44 notch filter (Thorlabs, CWL: 1064 nm, FWHM: 44 nm). The emission data complement the reflection information, providing a measure of the thermal state of the PZ and allowing an indirect assessment of the temperature of the heated and therefore emitting surface \cite{Taherkhani2021}. Both sensors had a measured field of view (FoV) of 3 mm in diameter, exceeding the typical melt pool size (100--300 \si{\micro\meter}), thus allowing monitoring of both the immediate melt area and its Heat-Affected Zone (HAZ).

\subsection{FPGA Board and Data Acquisition}

The novel aspect of our setup lies in its data acquisition and processing system, built around the Digilent Eclypse Z7 board with a Xilinx Zynq-7020 SoC. This system integrates a 667 MHz dual-core Cortex-A9 processor with an FPGA operating at 100 MHz. Two SYZYGY-compliant interfaces connect expansion modules: one with an AD9648 ADC (dual-channel, 14-bit, up to 125 MS/s) for recording photodiode signals at 100 kS/s, and another with an AD9717 DAC (dual-channel, 14-bit, 125 MS/s) for laser power modulation. The 100 kS/s sampling rate was chosen based on preliminary studies showing negligible frequency components above 50~kHz.

\subsection{Material}

The samples used were 2 mm thick plates of 316L stainless steel with a melting temperature of \SI{1400}{\celsius}. This material was selected due to its wide industrial applications, including use in chemical and petrochemical equipment \cite{Fajobi_2020}, food processing \cite{BARISH2013352}, and medical devices \cite{JAVIDI20081509}. Additionally, 316L's HAZ is easily recognizable in cross-sections due to its distinctive textural changes \cite{Rong2017}, making it ideal for post-processing analysis.

To test the algorithm's performance across a range of surface conditions, we experimented on three distinct sets of samples characterized by their surface roughness, quantified using the arithmetic mean height (\textit{Sa}). The first set consisted of brushed samples with \textit{Sa} = \SI{1.47}{\micro\meter}; the second set comprised sandblasted samples with \textit{Sa} = \SI{1.20}{\micro\meter}; and the third set included mixed samples with alternating brushed and sandblasted sections --- specifically, 10 mm brushed (\textit{Sa} = \SI{1.47}{\micro\meter}), 20 mm sandblasted (\textit{Sa} = \SI{1.23}{\micro\meter}), and 10 mm brushed (\textit{Sa} = \SI{1.47}{\micro\meter}).

This variety of surface conditions allowed us to evaluate the adaptability and robustness of our RL algorithm in different welding scenarios. All surface roughness measurements were performed using an S Neox 3D Optical Profiler (Sensofar Metrology, Spain) operating in white-light interferometry mode.

\section{Algorithm Implementation}

Our experimental setup employs an algorithm that integrates FPGA inference, data streaming to an external server, replay buffer updates, model training with Quantization-Aware Training (QAT), and continuous weight updates. This setup ensures that the FPGA is always operating with the most current model parameters, allowing rapid adaptability to changing conditions in the welding setup.

The process begins with the FPGA performing inference using pre-trained model weights stored in the Programmable Logic (PL) BRAM. During a single-line welding --- which we refer to as an episode --- data from the photodiodes are acquired through the ADC module at a sampling rate of 100~kS/s and processed by performing a forward pass of the neural network to compute control actions in real-time.

Upon the completion of an episode, the collected data --- including sensor signals and applied actions --- are transferred from the FPGA to the external server via Ethernet. This data transfer is managed by the ARM microcontroller on the System-on-Chip (SoC) --- also known as the processing system (PS). The server maintains a replay buffer, which stores experience tuples $(s,a,r,s')$ representing the state, action, reward, and next state for each time step. 

Between welding operations, the server performs model training using QAT. This process employs a digital twin of the FPGA created using the Python library Brevitas (Xilinx, USA) \cite{brevitas}, which simulates the FPGA's behavior, including quantization effects. After each training iteration, the server serializes the updated model weights and transmits them back to the SoC. The ARM microcontroller then uses Direct Memory Access (DMA) to update the PL BRAM with these new weights. 

The following pseudocode outlines the procedure of our experimental setup:

\begin{algorithmic}[1]
\Procedure{MainLoop}{}
    \State Initialize FPGA policy $\pi$ and digital twin
    \State Stream initial weights from server to PS
    \State Transfer weights from PS to PL BRAM via DMA
    \While{not converged}
        \State Acquire data from ADC on FPGA
        \If{OR signal $\geq 0.1V$}
            \For{$t = 1$ to $N_{\text{steps}}$}
                \State Acquire optical data $s_t$
                \State Sample $a_t \sim \mathcal{N}(\pi_\mu(s_t), \pi_\sigma(s_t))$
                \State Apply $a_t$ to laser
                \State Store $(s_t, a_t)$ pair in FIFO buffer
            \EndFor
        \EndIf
        \State Transfer data from FIFO buffer to PS via DMA
        \State Stream data to server via Ethernet
        \State Calculate $R_t = \text{OR}(s_{t+1}) \ \forall \ t$
        \State Perform $N_{\text{steps}}$ gradient steps of SAC learning
        \State Update model weights via QAT
        \State Transfer new weights to PS and to PL
    \EndWhile
\EndProcedure
\end{algorithmic}

\subsection{Reinforcement Learning Setup}
We formulate the laser welding process as a Partially Observable Markov Decision Process (POMDP) \cite{sutton2018reinforcement}, since the complicated dynamics of molten metal during welding are not directly observable. A POMDP is a discrete-time stochastic process in which at every time step $t$ the state $s_t$ represents the true condition of the weld, including the dynamics of the melting pool and the temperature distribution. An agent interacts with this environment by taking actions $a_t$, which in our case correspond to voltages controlling the laser power between 25W and 100W. However, due to the partial observability of the process, the agent must rely on observations $o_t$ as input. As noted before, our observations are derived from two optical sensors, providing indirect information about the melt pool dynamics. After each action, the environment transitions to a new state $s_{t+1}$, and the agent receives a reward $r(s_{t+1}, s_t, a_t)$ based on the quality of the weld.

The goal of the RL agent is to maximize the expected discounted return, given by $\mathbb{E}[\sum_{t=0}^T \gamma^t r(s_{t+1}, s_t, a_t)]$, where $\gamma$ is the discount factor. This approach optimizes for the sum of the given reward quantity for the entirety of the episode --- in this case one-line weld. Notice that, while reward engineering is challenging in general \cite{gupta2022unpacking}, choosing a good reward for laser welding is even more difficult, since many useful metrics of weld quality (such as weld depth used by \cite{gunther2016intelligent} or weld track width used by \cite{kaneko2023reinforcement}) are very challenging to obtain in real-time, or have to be acquired by extensive post-hoc manual labor.

To overcome these limitations, we used the findings of Wittemer et al. \cite{Wittemer2023}, who provide valuable information on the relationship between optical signals and weld pool dynamics. Their research reveals that the OR signal reaches a peak value just before the melt pool transitions from conduction mode to keyhole mode, corresponding to the formation of the largest stable weld. 

Based on this understanding --- while we present a system that can in principle be used with any reward --- we choose a simple reward function which encourages large stable welds while staying within conduction mode (avoiding vaporization of material), using the OR signal as the reward:
\begin{align}
    r(s_{t+1}) = \frac{\text{OR}(s_{t+1})}{10}.
\end{align}
This simple reward scales between 0 and 1 based on the photodiode sensor values (0-10 V).

Each episode consists of 80 steps, each 10 ms long. For the first 25 episodes, actions are sampled uniformly at random for exploration; subsequently, they are drawn from the learned policy. Every 10 episodes, a test episode is conducted in which actions are derived solely from the mean of the policy distribution $\pi_\mu(s_t)$, without sampling.

As the core RL algorithm we chose Soft Actor-Critic \cite{haarnoja2018soft} with adaptive entropy tuning \cite{haarnoja2018algorithms}, since maximum entropy RL algorithms have been found to perform well on continuous control tasks \cite{eysenbach2021maximum}.
We used the following hyperparameters for the SAC agent: hidden layer sizes [32, 64], ReLU activation, target entropy of -2, learning rate of $3 \times 10^{-4}$, batch size of 100, and a discount factor of 0.99. The agent performs 80 gradient steps at each training iteration, using mini-batches of size 100 sampled from the replay buffer.

\subsection{Agent Implementation in HLS}

To achieve the low-latency inference required for the RL agent, we implemented a neural network policy on the FPGA using High-Level Synthesis (HLS) \cite{xilinx_hls_ug998}, enabling efficient hardware realization directly from high-level algorithm descriptions.

The policy network consists of a Multi Layer Perceptron (MLP) with an input layer that receives data signals from the ADC, two hidden layers that perform weighted sums and activation functions, and an output layer that produces the final inference results. The HLS implementation optimizes the network for efficient parallel processing and pipelining, fundamental for high-speed inference operations. The core clock of the HLS module is set at 100~MHz via the internal clock divider. Specifically, we designed a pipelined architecture in which the processing of the current data window overlaps with the acquisition of the subsequent one. This overlap ensures continuous operation without idle cycles and minimizes latency in computing control actions.

In particular, during data acquisition, each processing window consists of exactly one million clock cycles, corresponding to a 10~ms time frame due to the decimation factor of 1,000 at a 100~kS/s sampling rate. After the acquisition phase, the processing of the data window is completed in just 354 clock cycles (3.54~\si{\micro\second}), significantly shorter than the interval between two consecutive data-points from the ADC (10~\si{\micro\second}). This rapid processing ensures that the computed control action is available almost instantaneously relative to the sampling period.

To balance precision with resource utilization, we employ integer arithmetic for weights and biases, maintaining them at 8-bit precision. However, to prevent overflow during computations, we implement a growing bitwidth strategy for activation values. This approach allows the bitwidth to increase progressively through the network layers, ensuring computational accuracy while optimizing resource usage.

In the output layer, the integer values are scaled and converted to floating points to compute the mean $\pi_\mu(s_t)$ and standard deviation $\pi_\sigma(s_t)$ for action sampling using the reparameterization trick \cite{kingma2013auto}. The scaling factor for this conversion is a parameter provided by the digital twin, ensuring consistency between the FPGA implementation and the training environment:
\begin{align}
    a_t = \pi_\mu(s_t) + \pi_\sigma(s_t) \cdot \varepsilon, \quad \varepsilon \sim \mathcal{N}(0, 1).
\end{align}
To avoid generating normally distributed numbers on the FPGA, $\epsilon$ are sampled on the server and streamed along with the weights, matching the number of steps in the environment. 

For the activation functions, we primarily use the Rectified Linear Unit (ReLU) throughout the network due to its efficient implementation on the FPGA. However, for action squashing, we approximate the hyperbolic tangent function (\texttt{tanh}) using a piecewise polynomial approximation optimized for the FPGA architecture. 

The combination of these optimizations results in an efficient and fast MLP implementation on the FPGA, capable of meeting the real-time requirements of the laser welding control system. 

\section{Experimental Results}\label{sec:results}

To evaluate the performance of our RL laser welding control system, we performed experiments on three types of 316L stainless steel samples: brushed, sandblasted, and mixed surfaces. For each sample type, we performed welding operations using both our learned policy and an optimal constant power strategy for comparison.

The optimal constant power for each surface type was determined through a grid search over the laser power range from 25W to 100 W. Specifically, we selected the power setting that yielded the highest average OR signal along the weld lines, correlating with the best weld quality in conduction mode achievable with traditional constant power welding.

\begin{figure}[t]
    \centering
    \includegraphics[width=0.45\textwidth]{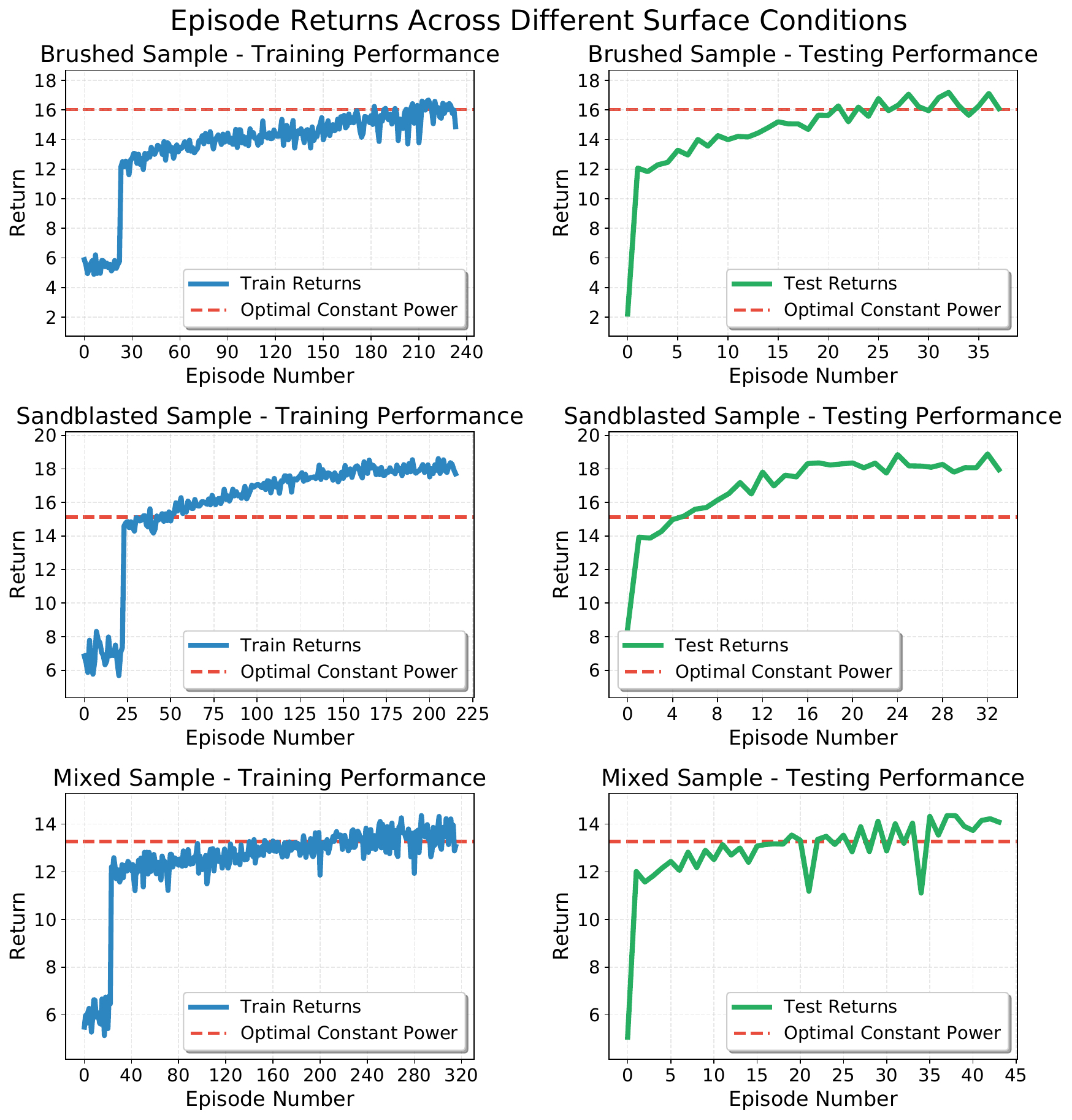}
    \vspace{-11pt}
    \caption{\footnotesize Comparison of our RL algorithm performance on different sample types during training and testing episodes. Three rows of plots are shown: Brushed, Sandblasted, and Mixed samples. Each row contains two graphs: Train Episode Returns (left) and Test Episode Returns (right). The blue lines represent the episode returns, while the red dashed lines indicate the Optimal Constant Power. }
    \label{fig:learning_curves}
    \vspace{-10pt}
\end{figure}

Figure \ref{fig:learning_curves} illustrates the learning progress and performance of our algorithm on the three sample types during both the training and testing episodes. The figure gives the episode return, which refers to the cumulative reward obtained by the agent over the course of a single episode. It is important to note the distinction between training and testing episodes. During training, the agent explores and exploits, leading to potentially noisy performance metrics. Test episodes, on the other hand, provide a cleaner and more stable measure of the agent's true capabilities by evaluating the learned policy's performance separately from the training process.

For brushed samples, the algorithm showed a gradual improvement in performance. The training curve indicates a slow but steady increase in return, eventually matching the optimal constant-power strategy after approximately 200 episodes. This gradual improvement reflects the challenge of outperforming an already optimized constant power on a uniform surface. Once stabilized, the policy achieved a 6.8\% higher test return than the baseline, demonstrating improvement even under seemingly optimal conditions.

In the case of sandblasted samples, our algorithm demonstrated its most impressive performance. The policy quickly learned to outperform the optimal constant power strategy, showing significant improvements within the first 40 episodes. This rapid learning and superior performance can be attributed to two key factors related to the nature of the sandblasted surfaces and the welding process itself. Firstly, sandblasting creates more homogeneous surfaces, which significantly reduces the variability in optical signals. This reduction in signal noise enhances the consistency of action selection and simplifies the learning process, as the policy's actions are derived directly from these optical inputs. Secondly, the welding process on sandblasted surfaces introduces a dynamic element that favors our adaptive approach. In fact, as welding begins, the sandblasting-induced surface roughness disappears. This transition creates a necessity for dynamic laser power adjustment, giving our policy an upper hand over the constant power baseline. The combination of these factors allowed our policy to achieve a substantial 22.6\% improvement over traditional optimized constant laser power strategies, highlighting the effectiveness of our approach in handling various welding scenarios.

\begin{figure}
\vspace{-4pt}
    \centering
    \includegraphics[width=0.32\textwidth]{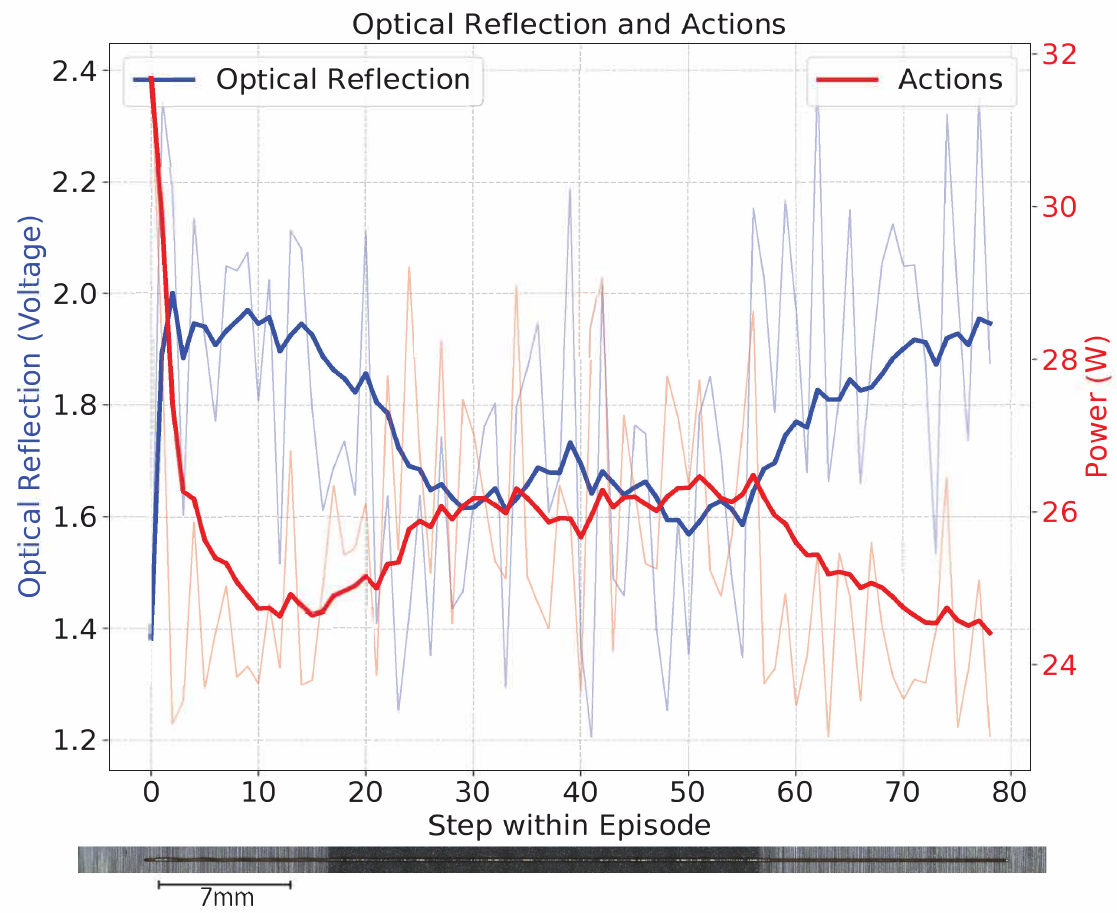}
    \vspace{-12pt}
    \caption{\footnotesize Comparison of OR and laser power actions during a test episode. OR (blue) and laser power actions (red). A power spike at the beginning initiates the melting process, while the increase in the middle of the episode corresponds to adjustments made for the sandblasted region. The microscope image at the bottom illustrates the corresponding processed line.}
    \label{fig:observation_vs_actions}
    \vspace{-20pt}
\end{figure}

The mixed sample scenario, which combined both brushed and sandblasted surfaces, presented a more complex challenge. In fact, the learning curve shows greater variability, reflecting the policy's attempts to adapt to changing surface conditions within single welding lines. After approximately 200 episodes, the algorithm successfully learned a policy that outperforms the optimal constant power approach, ultimately achieving a 7.30\% improvement in terms of test return. Figure \ref{fig:observation_vs_actions} illustrates the relationship between the OR and laser power actions during a test episode on this sample. Notably, the learned policy adopts an ideal power profile: it initiates the weld with a power spike to overcome the initial high reflectivity and start the melting process, then reduces the power to maintain stable welding on the brushed surface. Upon entering the sandblasted region (approximately steps 20 to 60), the policy increases the laser power to compensate for the surface condition change, as evidenced by the decrease in the OR signal. After exiting the sandblasted region, the policy reduces the power again when returning to the brushed surface. This dynamic adjustment demonstrates the policy's ability to respond effectively to varying surface conditions in real time.

\begin{figure}[htbp]
    \centering
    \includegraphics[width=0.4\textwidth]{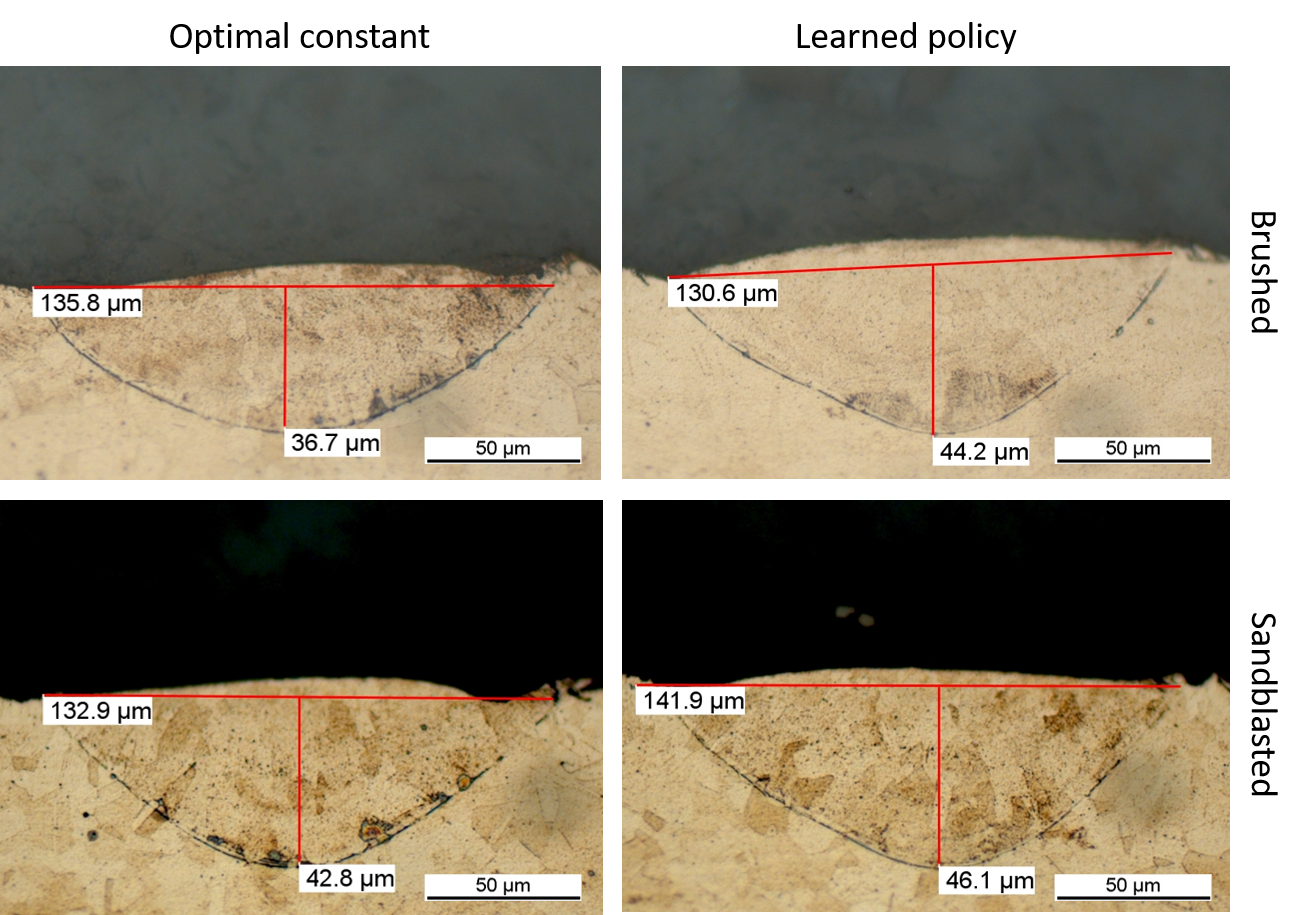}
    \caption{\footnotesize Representative melt-pool cross-sections taken from the mixed sample. \textbf{Left column:} Optimal constant power strategy. \textbf{Right column:} Learned policy. \textbf{Top row:} Brushed surface. \textbf{Bottom row:} Sandblasted surface. }
    \label{fig:melt_pool_cross_sections}
    \vspace{-20pt}
\end{figure}

To further validate our learned policy, we performed post-fabrication analysis of the welded samples. Figure~\ref{fig:melt_pool_cross_sections} compares representative melt-pool cross-sections obtained from the mixed sample, after grinding, polishing, and chemical etching with Aqua Regia, for both the optimal constant power strategy (left column) and our learned policy (right column) for both brushed (top row) and sandblasted (bottom row) surfaces. In each condition, the learned policy consistently produced wider and deeper melt pools while avoiding keyhole formation (that often leads to porosity \cite{Huang2022}). These enhanced melt pools indicate better penetration and fusion, which are critical for weld quality in conduction mode \cite{Fotovvati2018, Suder2024}. In this mode, porosity is minimal \cite{Barat2023}, leading to welds that typically exhibit more reliable mechanical properties. Indeed, tensile strength is strongly influenced by microstructural features and fusion zone characteristics \cite{Odermatt2021}, both of which correlate with melt-pool geometry. Accordingly, the larger melt pools observed here suggest improved tensile performance in addition to the apparent gains in penetration and fusion. While direct mechanical testing (e.g., tensile or fatigue tests) could offer further quantitative comparisons of weld performance, these cross-sectional results already underscore the policy’s adaptability and effective process optimization across diverse surface conditions.

\section{Discussion}\label{sec:discussion}
In this paper, we presented a methodology for generic laser material processing using RL, focusing on laser welding to showcase its feasibility. Our system demonstrates remarkable efficiency in both speed and resource utilization, successfully maintaining conduction-mode welding across various steel surfaces and identifying optimal power profiles with minimal training data. This sample- and time-effectiveness underlines its adaptability for rapid deployment in diverse manufacturing scenarios.

Nonetheless, real manufacturing conditions often involve factors not fully captured in our current setup. For instance, ambient temperature fluctuations or material additives can alter optical signals in ways that may require retraining or more robust state representations. Similarly, although basing the reward on the mean OR signal proved effective in avoiding keyhole formation, alternative reward functions encompassing a broader range of weld quality metrics --- such as porosity, tensile strength, or weld depth --- could further refine control.

Moreover, it is important to note that our current observation space, while sufficient for the presented experiments, may face challenges in more complex scenarios. Multi-material welding, for example, might generate similar sensor readings yet demand distinct power adjustments --- a phenomenon known as perceptual aliasing \cite{chrisman_reinforcement_learning_1992}. Additionally, at very high speeds, the strongest reflection may occur off-axis, rendering the coaxial signal maximization less effective. Future work could address such challenges by integrating advanced sensing technologies --- like Optical Emission Spectroscopy or off-axis measurements --- and by employing domain randomization \cite{tobin2017domain} or more diverse training data to accommodate environmental variability. In this way, the system would gain a richer understanding of the weld pool and be more robust to changes in both material and processing conditions.

Overall, the present results demonstrate how RL-driven laser welding can adaptively optimize process parameters in real time, capitalizing on FPGA-based low-latency inference to keep up with high-speed manufacturing demands. By refining the reward function and observation space, and by accounting for environmental factors, this approach has the potential to generalize to a wide range of laser applications --- ultimately offering a scalable, automated solution to ensure consistently high-quality outputs.

\section{Conclusion}\label{sec:conclusion}
Laser material processing and in particular laser welding is a critical technology in industries such as automotive manufacturing, aerospace engineering, and electronics assembly, prized for its precision, speed, and ability to produce high-quality welds. However, ensuring consistent weld quality in varying material properties remains a significant challenge, often requiring painstaking manual optimization.

In this paper, we have proposed a novel control strategy that addresses these limitations. Our approach employs a reinforcement learning algorithm implemented on an FPGA, achieving extremely low latency. This rapid processing allows the system to send control actions to the laser power modulation almost instantaneously, even before receiving the next signal data-point, ensuring real-time responsiveness. Furthermore, our method operates effectively without requiring extensive tuning of the reward function, making it highly adaptable to different welding scenarios.

Our experimental results demonstrated significant improvements over static power strategies in various surface conditions, including brushed, sandblasted, and mixed steel samples. The system showed remarkable adaptability, learning to outperform optimal constant power strategies by up to 22.58\% on sandblasted surfaces and 7.30\% on mixed surfaces in terms of the reward function. Although the reward serves as a useful proxy, we recognize that the ultimate quality of the weld is determined by physical metrics. To this end, we conducted post-process analyses of the welded samples, including examining cross-sectional images. These cross sections revealed that our learned policy consistently produced larger and deeper melt pools compared to the constant power strategy, indicating improved weld penetration and overall quality.

In conclusion, our real-time, adaptive laser welding control system represents a significant step forward in automating and optimizing laser welding processes, and its generality makes it able to target many other laser applications with potential benefits across a wide range of industries.






\section*{Acknowledgement}

The authors would like to express their gratitude to Thomas Rytz, Toni Ivas, Peter Ramseier, and Pinku Yadav for their invaluable support and contributions to this research. We also extend our thanks to the Strategic Focus Area (SFA) Advanced Manufacturing program for financing this project.

\bibliographystyle{IEEEtran}
\bibliography{IEEEabrv,refs}


\end{document}